\documentclass[conference]{IEEEtran}
\pdfoutput=1
\IEEEoverridecommandlockouts
\usepackage{cite}
\usepackage[utf8]{inputenc}
\usepackage{amsfonts}

\usepackage{graphicx}
\usepackage{textcomp}
\usepackage{xcolor}
\usepackage{lipsum}
\usepackage[ruled,vlined]{algorithm2e}
\usepackage{enumerate}
\usepackage{tabularx,booktabs}

\usepackage{amsmath}
\usepackage{amssymb}

\usepackage{subfigure}
\usepackage{dsfont}
\usepackage{diagbox}
\usepackage[hidelinks]{hyperref}
\usepackage{multicol}
\usepackage{bm}
\usepackage{float}
\restylefloat{table}
\renewcommand{\figurename}{Figure}
\usepackage{fancyhdr} 
\def\BibTeX{{\rm B\kern-.05em{\sc i\kern-.025em b}\kern-.08em
    T\kern-.1667em\lower.7ex\hbox{E}\kern-.125emX}}

\makeatletter
\newcommand{\removelatexerror}{\let\@latex@error\@gobble}
\makeatother

\begin{document}
\pagestyle{fancy}
\title{Neighborhood Random Walk Graph Sampling for Regularized Bayesian Graph Convolutional Neural Networks}

\author{
    \IEEEauthorblockN{Aneesh Komanduri and Justin Zhan}
    \IEEEauthorblockA{
        \textit{Department of Computer Science and Computer Engineering} \\
        \textit{University of Arkansas}, Fayetteville, Arkansas, USA \\
        \{akomandu, jzhan\}@uark.edu
    }
}

\maketitle

\begin{abstract}
In the modern age of social media and networks, graph representations of real-world phenomena have become an incredibly useful source to mine insights. Often, we are interested in understanding how entities in a graph are interconnected. The Graph Neural Network (GNN) has proven to be a very useful tool in a variety of graph learning tasks including node classification, link prediction, and edge classification. However, in most of these tasks, the graph data we are working with may be noisy and may contain spurious edges. That is, there is a lot of uncertainty associated with the underlying graph structure. Recent approaches to modeling uncertainty have been to use a Bayesian framework and view the graph as a random variable with probabilities associated with model parameters. Introducing the Bayesian paradigm to graph-based models, specifically for semi-supervised node classification, has been shown to yield higher classification accuracies. However, the method of graph inference proposed in recent work does not take into account the structure of the graph. In this paper, we propose a novel algorithm called Bayesian Graph Convolutional Network using Neighborhood Random Walk Sampling (BGCN-NRWS), which uses a Markov Chain Monte Carlo (MCMC) based graph sampling algorithm utilizing graph structure, reduces overfitting by using a variational inference layer, and yields consistently competitive classification results compared to the state-of-the-art in semi-supervised node classification.
\end{abstract}

\begin{IEEEkeywords}
Semi-Supervised Learning, Graph Convolutional Neural Networks, Bayesian Inference, Node Classification, Markov Chain Monte Carlo
\end{IEEEkeywords}

\section{Introduction}

%
%



\IEEEPARstart{G}RAPH Neural Networks (GNNs), proposed by Gori et al, \cite{4700287} have become a useful method for predicting certain behaviors in graphs. Examples include node classification, edge classification, and recommendation systems \cite{inproceedings}, among others. This paper focuses on the node classification task. This is a classic classification problem where we are given an observed graph with partially labeled nodes and our task is to train our model on this data so that given a node and its features, the model will be able to predict the correct label of an unlabeled node up to some degree of accuracy. This is an incredibly important task because, in the real world, true labels are expensive to obtain, so we rely on techniques such as random sampling to get these labels and use these labels to develop label prediction models. 


To improve on classification accuracies from the GNN, many convolution-based approaches such as the spectral convolution \cite{bruna2014spectral}, deep convolution \cite{henaff2015deep}, and \cite{duvenaud2015convolutional} were proposed for the graph data structure. Defferrard et al \cite{defferrard2017convolutional} proposed the Graph Convolutional Neural Network (GCN) and \cite{kipf2017semisupervised} proposed an efficient version of Graph Convolutional Networks for semi-supervised learning and will be the base model we use for model comparison. Due to the fact that convolution operations are spatial operations, meaning that they take into account contextual information, it helps a node in a graph to find its place in the graph with respect to its neighboring nodes. This method has proven to be among the best performing architectures for a variety of graph learning tasks. However, GCNs fail to account for the uncertainty in the underlying structure of a graph such as noisy data with spurious edges. Further, since the graph convolution in the GCN is a special case of Laplacian smoothing, it can cause model overfitting and over-smoothing so that nodes in the graph may be indistinguishable.

The majority of the existing approaches process the graph as the ground truth. However, in many practical settings, the graph is often constructed from noisy data or invalid modeling assumptions. As a result, there may be many spurious edges, or edges between very similar nodes might be omitted. This can lead to a decrease in the performance of learning algorithms. Various existing approaches such as
the graph attention network \cite{velickovic2018graph}, graph ensemble based approach \cite{anirudh2018bootstrapping}, multiple adjacency matrices \cite{Such_2017}, dual graphs \cite{monti2018dualprimal}, and skip connections \cite{sukhbaatar2016learning} address the performance issue partially. However, none of these methods have the flexibility to add edges that could be missing from the observed graph.

We can tackle this issue by simply viewing the observed graph as a sample from a parametric random variable and target joint inference of the graph and GCN weights using a Bayesian scheme called a Bayesian Graph Convolutional Neural Network (BGCN) \cite{zhang2018bayesian}. The framework of a BGCN allows graphs to be sampled randomly, which helps to account for uncertainty in the graph thereby increasing accuracy in node classification. 

 Pal et al \cite{pal2019bayesian} use a scheme called node copying, similar to \cite{chang}, to sample graphs from the observed graph under some probability distribution with a fixed hyperparameter and non-parametric graph learning approaches \cite{pal2019bayesian}. However, we show that a more effective technique is to use the notion of rejection sampling and Markov Chain Monte Carlo methods such as Metropolis-Hastings \cite{7113345} to sample graphs based on graph structure. Further, we propose an addition to Dropout regularization\cite{JMLR:v15:srivastava14a} to decrease overfitting.

Our contributions are three-fold. Firstly, we propose a Markov Chain Monte Carlo (MCMC) random walk graph sampling algorithm \cite{akomand} to utilize graph structure to perform sampling and increase distinguishability between nodes. Secondly, we use a variation inference linear model in addition to the base GCN as a regularizer for graph convolutions to reduce overfitting. This is performed by using Bayes by Backprop to regularize parameters between graph convolutions. Lastly, we evaluate BGCN-NRWS in semi-supervised node classification on three benchmark datasets: Cora, Citeseer, and Pubmed. Experimental results reveal that BGCN-NRWS achieves new state-of-the-art results with varying degrees of supervision.


The paper is organized as follows. We discuss some background pertaining to GNNs and Bayesian inference in Section II. We present the workings of the BGCN and introduce the Metropolis-Hastings neighborhood random walk graph sampling (NRWS) variant in Section III. Section IV describes the datasets used and outlines the experimentation procedure and summarizes the results. Finally, we conclude the paper and discuss potential future work in Section V.

\section{Background}
\label{sec:background}

\subsection{Graph Convolutional Networks}
 The Graph Convolutional Neural Network, proposed by Defferrard et al \cite{defferrard2017convolutional} and Kipf et al \cite{kipf2017semisupervised}, is a variant of the GNN. We define the observed graph as $\mathcal{G}_{obs} = (\mathcal{V}, \mathcal{E})$, where $\mathcal{V}$ is the set of $N$ nodes and $\mathcal{E} \subseteq \mathcal{V} \times \mathcal{V}$ is the set of edges. Let $\mathbf{X} = [\bm{x_1}, \bm{x_2}, \dots, \bm{x_N}]^T \in \mathbb{R}^{d\times N}$ be the feature matrix, where $x_i$ denotes the $i$th node's $d$-dimensional feature vector and $\mathbf{Y}_{\mathcal{L}} = \{\mathbf{y}_i \mid i\in \mathcal{L}\}$, where $\mathbf{y}_i$ denotes the $i$th node's label and $\mathcal{L}\subset \mathcal{V}$ is the number of node labels known for the training set. The goal is to predict the labels of the remaining nodes in the observed graph.



The graph convolution operation takes the input graph, propagates through hidden layers, and flattens the representation by aggregating features and ultimately condensing it down to a layer with labels. Once this is done, a simple softmax can be taken to determine the probabilities of each label. 

The Propagation Rule of the GCN is as follows:

\begin{equation}
    \mathbf{H}^{(1)} = \sigma(\mathbf{\Tilde{D}}^{-\frac{1}{2}}\mathbf{\Tilde{A}}\mathbf{\Tilde{D}}^{-\frac{1}{2}}\mathbf{X}\mathbf{W}^{(0)})
\end{equation}
\begin{equation}
    \mathbf{H}^{(l+1)} = \sigma(\mathbf{\Tilde{D}}^{-\frac{1}{2}}\mathbf{\Tilde{A}}\mathbf{\Tilde{D}}^{-\frac{1}{2}}\mathbf{H}^{(l)}\mathbf{W}^{(l)})
\end{equation}
where $\mathbf{H}^{(l+1)}$ represents the hidden features (activations) at the next hidden layer, $\mathbf{\Tilde{A}}$ is the adjacency matrix with self-loops accounted for so that each node includes its own features at its next representation, and $\mathbf{\Tilde{D}}$ is the Degree Matrix of $\mathbf{\Tilde{A}}$ which is used to normalize nodes with large degrees because otherwise, nodes with a large number of neighbors can be computationally demanding. We have that $(\mathbf{\Tilde{D}^{-\frac{1}{2}}\Tilde{A}\Tilde{D}^{-\frac{1}{2}}})\in \mathbb{R}^{n\times n}$, $\mathbf{H}^{(l)}\in \mathbb{R}^{n\times d}$, and $\mathbf{W}^{(l)}\in \mathbb{R}^{d\times n}$ \cite{kipf2017semisupervised}.
In a 2D image convolution, we have aggregation of pixels in a neighborhood to perform convolution and reduce the spatial dimension, whereas a graph convolution requires a similar aggregation, except it is the aggregation of the number of neighboring nodes that may vary.


\subsection{Bayesian Neural Networks}
A common definition of a Bayesian neural network is a stochastic artificial neural network that is trained using Bayesian inference \cite{jospin2020handson}. Generally, in regular neural networks, the network weights are not treated as random variables. Rather, they are assumed to have a true value that may not be known yet. Bayesian Neural Networks (BNNs) treat weights as a random variable and learn the model weights based on the information we have at hand. The goal is to learn a distribution of the weights (or any relevant parameters) conditional on what we observe in the data \cite{Goan_2020}. 






Now, we can apply this process to a supervised learning classification setting. Let $\mathbf{X} = \{x_1,\dots, x_n\}$ be the training inputs and $\mathbf{Y} = \{y_1,\dots, y_n\}$ be the corresponding outputs. The goal is to learn a function $y = f(x)$ using a neural network to find a relationship between $x$ and $y$. Using the Bayesian framework, we model the weights of our network $\mathcal{W}$ as a random variable with a prior distribution introduced over them. The weights are not deterministic parameters, so the outputs of the neural network will also be random variables. So, we compute the marginal likelihood of the outputs conditioned on the set of inputs and outputs as follows

\begin{equation}
    p(y|x, \mathbf{X}, \mathbf{Y}) = \int p(y|x, \mathbf{W})p(\mathbf{W}| \mathbf{X}, \mathbf{Y}) \; d\mathbf{W}
\end{equation}

where $p(y|x, \mathbf{W})$ is the likelihood that can be computed by applying a softmax function to the output of the network. This integral is intractable, so we can use Monte Carlo techniques, which are discussed later in \hyperref[sec:MC]{Section 4.2}, to approximate the integral. Namely, we can approximate the posterior as follows:

\begin{equation}
    p(y|x, \mathbf{X}, \mathbf{Y}) \approx \frac{1}{T}\sum_{i=1}^S p(y|x, \mathbf{W}_i)
\end{equation}

where $T$ is the number of samples drawn, $S$ is the number of weights $\mathbf{W}_i$ obtained using Monte Carlo dropout \cite{gal2016dropout}.

\section{Approach}
\label{sec:approach}
Generally in the Bayesian paradigm, as we described in the graph setting, the observed graph is viewed as a random variable and we are trying to use Bayesian methods to infer the posterior for the underlying graph \cite{gal2016bayesian} \cite{pal2019bayesian}. Based on this approach, we propose a graph sampling method that can be effective on three fronts. Our approach: utilizes graph structure to determine acceptance probability during sampling, proposes an MCMC random walk-based algorithm that allows for diverse connections, and allows the observed graph to learn connections between weakly linked nodes and potentially missing connections. To accomplish this, we propose a unique MCMC based random walk technique to sample nodes from the graph, which gives us an approximation to a Bayesian GNN as stated in the following Corollary.

\bigskip

\newtheorem{corollary}{\normalfont \bf{Corollary}}

\begin{corollary} \normalfont
Any graph neural network with random walk
sampling is an approximation of a
Bayesian graph neural network as long as outputs are calculated using Monte Carlo sampling \cite{hasanzadeh2020bayesian}.
\end{corollary}

\bigskip

A previous method proposed in \cite{pal2019bayesian} utilizes a version of node copying for graph inference, which copies the neighbors of a node with the same label as the current node to the current node's neighbors. The problem with this approach is that high node densities can lead to overfitting of the GCN when applied to the graph. Additionally, unlike our approach, the node copying approach uses a fixed hyperparameter $\epsilon$ to determine the acceptance probability, which does not use the structure of the graph in any form. The arbitrary nature of this proposal can prove to be problematic when generalizing to larger graph datasets. 


\subsection{Neighborhood Random Walk Sampling}
Our approach shown in Algorithm \ref{algorithm3} utilizes the Neighborhood Random Walk Sampling method to sample nodes from the observed graph and copy neighborhoods of nodes from the random walk. In order to sample graph $\mathcal{G}$ from the proposed model, we utilize the Metropolis-Hastings random walk. Recall the Metropolis-Hastings algorithm from the previous section. We will reformulate this algorithm and apply it to graph sampling. 

Suppose that we want to generate a random variable $\mathbf{V}$ taking values $\{1, 2, \dots, n\}$, representing nodes in our graph, according to some target distribution $\{\pi_i\}$ where $i\in \mathbf{V}$. We have that $\pi_i = \frac{b_i}{\beta}$ where $b_i > 0$ and the normalizing constant $\beta$ is difficult to estimate. We simulate a markov chain such that the stationary distribution of the chain will converge to the target distribution of the posterior that we have sought after. Let $\{\mathbf{X}_t \mid t=0,1,\dots\}$ be the state-space of the markov chain $\mathbf{M}$ using a transition probability matrix $\mathbf{Q}$, which we will assume is the proposal distribution for our method. Now, if $\mathbf{X}_t = i$, then we generate a candidate sample $\mathbf{Y}$ such that $P(\mathbf{Y} = j) = \mathbf{q}_{ij}$ for all $i,j\in V$. Then, we have the following acceptance function:

\begin{equation}
    \alpha(j \mid i) = \min \Big\{1, \frac{\mathbf{q}(j\mid i)\pi_j}{\mathbf{q}(i\mid j)\pi_i}\Big\}
\end{equation}

Now, to accomplish graph sampling, we treat the nodes in the observed graph $\mathcal{G}$ as states in a Markov chain. We uniformly sample a neighbor of node $i$ from the proposal distribution $\mathbf{Q}$, say $j$. Since we are going to perform a random walk, the transition probability of moving from state (or node) $i$ to state $j$ is simply $1/d_i$, the degree of the current node $i$. The transition matrix $\mathbf{Q}$ is a stochastic symmetric positive semi-definite matrix and therefore is a symmetric distribution. For a uniform target, we have that $\pi_i = \pi_j$. Further, $\mathbf{q}(j\mid i) = 1/d_j$ and $\mathbf{q}(i\mid j) = 1/d_i$. So, our new acceptance function for sampling from graphs is
\begin{equation}
    \alpha(j \mid i) = \min \Big\{1, \frac{d_i}{d_j}\Big\}
\end{equation}

Let $\xi \in \{1,2,\dots, n\}^n$ be a random vector where $n$ denotes the total number of nodes and the $j$th entry $\xi^{j}$ denotes the node whose edges are to be copied to the $j$th node in the observed graph $\mathcal{G}_{obs}$ \cite{pal2019bayesian}. This random vector is similar to the random vector used for the node copying scheme in Pal et al.

\begin{figure}
    \centering
    \removelatexerror
\begin{algorithm}[H]
        \label{algorithm3}
        \SetAlgoLined
        \textbf{Input:} $\mathcal{G}_{obs}$, $\mathbf{X}$, $\mathbf{Y}_{\mathcal{L}}$ 
        
        \textbf{Output:} $p(\mathbf{Z}\mid \mathbf{Y}_{\mathcal{L}}, \mathbf{X}, \mathcal{G}_{obs})$ 
        
         Initialization: Execute a pre-training step for a classifier to obtain predicted labels \\
         \For{i = 1 \KwTo V} {
            Sample $\xi_v \sim p(\xi \mid \mathcal{G}_{obs}, \mathbf{X}, \mathbf{Y}_{\mathcal{L}})$ 
            
            \For{k = 1 \KwTo $N_{G}$} {
                    \upshape{Sample $j\in \mathcal{N}(i)$} 
                    
                    \upshape{Uniformly sample $u \sim U(0, 1)$} 
                    
                    \eIf{$u \leq \min\{1, d_i/d_j\}$} {
                        \upshape{accept node $j$}
                    }{
                        \upshape{reject $j$ and stay at node $i$}
                    }

                \upshape{Copy neighborhood of accepted node to the neighbors of candidate node}
            }
         } 
         {
         \text{Run BGCN regularized classifier with inferred graph}
         } 
         \caption{Neighborhood Random Walk Sampling}
\end{algorithm}
\end{figure}

Let $\mathbf{X}\in \mathbb{R}^{n\times d}$ represent the node features and $\mathbf{Y}_L \in \mathbb{R}^n$ represent the training labels.  
That is, each row of $\mathbf{X}$ is the $d$-dimensional feature vector of the corresponding node. We want to predict one of $K$ class labels $\hat{c}_m\in \{1, 2, \dots, K\}$ for each node $m$ in the graph. Define the posterior distribution of the random vector $\xi$ as follows:

\begin{equation}
    p(\xi \mid \mathcal{G}_{obs}, \mathbf{X}, \mathbf{Y}_L) = \prod_{j=1}^n p(\xi^j\mid \mathcal{G}_{obs}, \mathbf{X}, \mathbf{Y}_L)
\end{equation}

\begin{equation}
 p(\xi^j = i \mid \mathcal{G}_{obs}, \mathbf{X}, \mathbf{Y}_L) = \begin{cases}
     1/d_{i}, & \text{ if $i\in \mathcal{N}(j)$} \\
     0, & \text{otherwise}
 \end{cases}
\end{equation}

Sampling a node $\xi^j$ is done by selecting a node at random from a random walk generated by a Markov chain and copying the neighbors of that node. As stated before, the nodes in the observed graph can be viewed as states of the Markov chain. A random walk is generated from the current node and a random neighbor of the terminating node is chosen as the candidate to sample. This candidate is accepted based on an acceptance function calculated based on the ratio of the degree of the current node and the degree of the candidate node. The sampling is carried out by simply copying the $\xi^j$'th node of $\mathcal{G}_{obs}$ in the place of the $j$th node of $\mathcal{G}$ independently for all $1\leq j \leq n$ with probability given by the acceptance function. 
\\


Now, assuming that the events (the states), namely $\xi$, are independently and identically distributed, our generative model is as follows:

\begin{align}
    p(\mathcal{G}| \mathcal{G}_{obs}, \xi) =  \prod_{j=1}^n \gamma^{\mathds{1}_{\mathcal{G}_j= \mathcal{G}_{obs,j}}} \Big(1 - \gamma\Big)^{\mathds{1}_{\mathcal{G}_j= \mathcal{G}_{obs,\xi^j}}}
\end{align}

where $\mathds{1}_{\mathcal{G}_q = \mathcal{G}_{obs,j}}$ is the indicator function of copying node $q$ in the observed graph $\mathcal{G}_{obs}$ in place of node $j$ of sampled graph $\mathcal{G}$ and $\gamma = \min \Big\{1, \frac{d_i}{d_j}\Big\}$. This model changes the neighbors of the $j$th node with the neighbors of the node reached by the random walk.

\subsection{Variational Inference Layer}
In addition to the random walk sampling mechanism used to approximate a Bayesian GNN, we develop a modified Bayes-by-Backprop layer that takes into account the feature matrix and adjacency matrix of the graph to act as a self regularization layer between graph convolutions \cite{blundell2015weight}. Variational Inference is a deterministic approach that serves as a fix to a model suffering from overfitting. That is, weights in the network are probability distributions and we are trying to approximate the posterior distributions. As mentioned before, we use Monte Carlo dropout regularization to prevent overfitting, however variational inference automatically leads to regularization, thereby inducing a more accurate model for our data. 

With Bayesian neural networks, we are trying to approximate the posterior quantity $P(\mathbf{W}|\mathbf{X})$ by finding parameters $\Theta$ of a distribution on the weights $q(\mathbf{W} | \Theta)$, which is the variational posterior, that minimize the KL-divergence with the true posterior quantity shown as follows:

\begin{align*}
    \Theta^* = \min_{\Theta} \text{KL}[q(\mathbf{W}\mid \Theta) || P(\mathbf{W})] - \mathbb{E}_{q(\mathbf{W}|\Theta)} [\log P(\mathbf{X}\mid \mathbf{W})]
\end{align*}

We can approximate the parameter distribution using Monte Carlo sampling as follows:

\begin{equation}
    \Theta^* = \sum_{i=1}^n \log q(\mathbf{W}^{(i)}|\Theta) - \log P(\mathbf{W}^{(i)}) - \log P(X|\mathbf{W}^{(i)})
\end{equation}

As in \cite{blundell2015weight}, we introduce a Gaussian density as the prior $P(\mathbf{W})$ over the weights as follows:

\begin{equation}
    P(\mathbf{W}) = \prod_{i} \mathcal{N}(\mathbf{W}_i \mid 0, \sigma_p^2)
\end{equation}

where $\mathcal{W}_i$ is the $i$th weight of the network, the distribution is centered at mean 0, and $\sigma_p^2$ is the variance of the Gaussian density, which is set to $0.1$ for our model. So, our variational posterior with a Gaussian distribution is defined as follows:

\begin{equation}
    q(\mathbf{W}\mid \Theta) = \prod_i \mathcal{N}(\mathbf{W}_i \mid \mu, \sigma^2)
\end{equation}

So, the combined loss function for our model is defined by:
\begin{equation}
    \mathcal{L}(\mathbf{X}, \Theta) = (\log q(\mathbf{W}\mid \Theta) - \log P(\mathbf{W})) - \log P(\mathbf{X} \mid \mathbf{W})
\end{equation}
The variational inference layer acts as a Bayesian regularization between graph convolutions to prevent overfitting and to effectively model uncertainty.

\subsection{Bayesian Graph Convolutional Networks}
Given that $\mathbf{Z}$ is the final output collected from the last layer of the network (the prediction), we compute the posterior probability of the node labels by marginalizing with respect to the graph and the GCN weights \cite{zhang2018bayesian}.

\begin{align*}
        p(\mathbf{Z}| \mathbf{Y}_{L}, \mathbf{X}, \mathcal{G}_{obs}) &= \int p(\mathbf{Z}| \mathbf{W}, \mathcal{G}_{obs}, \mathbf{X})p(\mathbf{W}| \mathbf{Y}_L, \mathbf{X}, \mathcal{G}) \\ &\qquad p(\mathcal{G}| \mathcal{G}_{obs}, \xi)p(\xi| \mathcal{G}_{obs}, \mathbf{Y}_L, \mathbf{X})\; d\mathbf{W} \;d\mathcal{G}\; d\xi
\end{align*}

where $\mathcal{W}$ is the random weight matrix of the BGCN over the graph $\mathcal{G}$ and $\xi$ is an $n$-dimensional random vector that represents the neighborhood random walk sampling model. Simply put, this continuous probability consists of the softmax probability, prior distribution on the weights, generative model for graph sampling, and our model for the posterior for the random variable $\xi$. Similar to Pal et al., our approach models the marginal posterior distribution of the graph $\mathcal{G}$ as $p(\mathcal{G}\mid \mathcal{G}_{obs}, \mathbf{X}, \mathbf{Y}_L)$. This allows the features $\mathbf{X}$ to play a role in the graph inference process. This posterior is intractable, so Monte Carlo techniques are required to approximate the value.
\begin{equation*}
    p(\mathbf{Z}| \mathbf{Y}_{L}, \mathbf{X}, \mathcal{G}_{obs}) \approx \frac{1}{V}\sum_{v=1}^V\frac{1}{N_GS}\sum_{i=1}^{N_G}\sum_{s=1}^S p(\mathbf{Z}| \mathbf{W}_{s,i,v}, \mathcal{G}_{obs}, \mathbf{X})
\end{equation*}

where $V$ samples $\xi_v$ are drawn from the $p(\xi | \mathcal{G}_{obs}, \mathbf{Y}_L, \mathbf{X})$ distribution. The $N_G$ sampled graphs are sampled from $p(\mathcal{G} | \mathcal{G}_{obs}, \xi_v)$ and $S$ weight matrices are sampled from $p(\mathbf{W}| \mathbf{Y}_L, \mathbf{X}, \mathcal{G}_{i,v})$ from the Bayesian GCN corresponding to the sampled graph. The complete algorithm is outlined in Algorithm \ref{algorithm3}.

The GCN weights for Bayesian inference can be obtained by performing a variety of techniques such as expectation propagation \cite{hernandezlobato2015probabilistic}, variational inference \cite{gal2016dropout} \cite{pmlr-v54-sun17b} \cite{louizos2017multiplicative}, and Markov chain Monte Carlo methods \cite{Neal}. We utilize the Bayes by Backprop variational inference introduced by Blundell et al \cite{blundell2015weight} as a regularizer linear layer between graph convolutions.
We have found that our neighborhood random walk sampling is quite effective for a multitude of reasons. Firstly, a flaw of the node copying scheme proposed by Pal et al is that the sampling occurs under a fixed probability parameter $\epsilon$. Instead of fixing a parameter, our model uses the structure of the graph and an acceptance function to perform rejection sampling. Secondly, the node copying scheme copies the neighborhood of nodes that have the same label as the current node, which can lead to overfitting of the model. Nodes can be overly dense, which can cause the model classification accuracy to suffer. Rather, performing a Markov chain-based random walk throughout the graph over a number of iterations (set as a hyperparameter) can be beneficial for sampling efficiency. Many connections in a noisy graph can be loose connections. A random walk can help diversify connection to the current node thereby broadening information when aggregated together from neighbors when being propagated through the graph neural network. Lastly, our proposed model has $\mathcal{O}(N)$ complexity, which is much more efficient than the Mixed Membership Stochastic Block Model sampling method \cite{zhang2018bayesian}, which has complexity $\mathcal{O}(N^2)$.

\section{Experiments \& Results}
\label{sec:evaluation}
We performed multiple experiments to validate our findings. Specifically, our model increases the node classification accuracy in a graph. We compare the performance of our BGCN-NRWS model with that of CheybyNet\cite{defferrard2017convolutional}, GCN\cite{kipf2017semisupervised}, GAT\cite{velickovic2018graph}, BGCN \cite{zhang2018bayesian}, and BGCN using Node Copying (BGCN-NC) \cite{pal2019bayesian} for the node classification task. Experiments were carried out on an NVIDIA 2x Quadro RTX 8000 GPU.
\subsection{Datasets}
We evaluate the performance of the proposed Bayesian
GCN on three well-known citation datasets: Cora, CiteSeer, and Pubmed \cite{Sen_Namata_Bilgic_Getoor_Galligher_Eliassi-Rad_2008}. In these
datasets, each node represents a document and has a sparse
bag-of-words feature vector associated with it. Edges are
formed whenever one document cites another. The direction
of the citation is ignored and an undirected graph with a
symmetric adjacency matrix is constructed. Each node label represents the topic that is associated with the document.
We assume that we have access to several labels per class
and the goal is to predict the unknown document labels.

\textbf{\underline{Cora}}: The Cora dataset consists of 2708 scientific publications classified into one of 7 classes. The citation network consists of 5429 links. Each publication in the dataset is described by a 0/1-valued word vector indicating the absence/presence of the corresponding word from the dictionary. The dictionary consists of 1433 unique words.

\textbf{\underline{Citeseer}}: The CiteSeer dataset consists of 3312 scientific publications classified into one of 6 classes. The citation network consists of 4732 links. Each publication in the dataset is described by a 0/1-valued word vector indicating the absence/presence of the corresponding word from the dictionary. The dictionary consists of 3703 unique words.

\textbf{\underline{Pubmed}}: The Pubmed Diabetes dataset consists of 19717 scientific publications from the PubMed database pertaining to diabetes classified into one of 3 classes. The citation network consists of 44338 links. Each publication in the dataset is described by a TF/IDF weighted word vector from a dictionary that consists of 500 unique words.

\begin{center}
\begin{table}[H]
 \begin{tabular}{c | c c c} 
 \hline
 \textbf{Random Split} & \textbf{5 labels} & \textbf{10 labels} & \textbf{20 labels} \\ [0.5ex] 
 \hline
 \textbf{ChebyNet}   & \centering{$61.7 \pm 6.8$}    & \centering{$72.5 \pm 3.5$} & $78.8 \pm 1.6$ \\
 \textbf{GCN} & \centering{$70.0 \pm 3.7$}    & \centering{$76.0 \pm 2.2$} &  $79.8 \pm 1.8$ \\
 \textbf{GAT} & \centering{$70.4 \pm 3.7$}    & \centering{$76.6 \pm 2.8$} & $79.9 \pm 1.8$ \\
 \textbf{BGCN-NC} & \centering{$73.8 \pm 2.7$}    & \centering{$77.6 \pm 2.6$} & $80.3 \pm 1.6$ \\
 \textbf{BGCN-NRWS} & \centering{$\bf{79.5 \pm 2.8}$}    & \centering{$\bf{82.2 \pm 1.4}$} & $\bf{83.3 \pm 1.3}$ \\
 \hline
\end{tabular}
\vspace{0.2cm}
    \caption{Classification accuracy for Cora dataset}
    \label{tab:t1}
\end{table}
\end{center}

\vspace{-1cm}

\begin{center}
\begin{table}[H]
 \begin{tabular}{c | c c c} 
 \hline
 \textbf{Random Split} & \textbf{5 labels} & \textbf{10 labels} & \textbf{20 labels} \\ [0.5ex] 
 \hline
 
 \textbf{ChebyNet}   & \centering{$58.5 \pm 4.8$}    & \centering{$65.8 \pm 2.8$} &  $67.5 \pm 1.9$ \\
 
 \textbf{GCN} & \centering{$58.5 \pm 4.7$}    & \centering{$65.4 \pm 2.6$} &  $67.8 \pm 2.3$ \\
 
 \textbf{GAT} & \centering{$56.7 \pm 5.1$}    & \centering{$64.1 \pm 3.3$} &  $67.6 \pm 2.3$ \\
 
 \textbf{BGCN-NC} & \centering{$63.9 \pm 4.2$}    & \centering{$68.5 \pm 2.3$} &  $70.2 \pm 2.0$ \\
 
 \textbf{BGCN-NRWS} & \centering{$\bf{66.8 \pm 2.8}$}    & \centering{$\bf{71.2 \pm 2.8}$} &  $\bf{72.5 \pm 2.8}$ \\
 \hline
\end{tabular}
\vspace{0.2cm}
    \caption{Classification accuracy for Citeseer dataset}
    \label{tab:t2}
\end{table}
\end{center}

\vspace{-1cm}

\begin{center}
\begin{table}[H]
 \begin{tabular}{c | c c c} 
 \hline
 \textbf{Random Split} & \textbf{5 labels} & \textbf{10 labels} & \textbf{20 labels} \\ [0.5ex] 
 \hline
 
 \textbf{ChebyNet}   & \centering{$62.7 \pm 6.9$}    & \centering{$68.6 \pm 5.0$} &  $74.3 \pm 3.0$ \\
 
 \textbf{GCN} & \centering{$69.7 \pm 4.5$}    & \centering{$73.9 \pm 3.4$} &  $77.5 \pm 2.5$ \\
 
 \textbf{GAT} & \centering{$68.0 \pm 4.8$}    & \centering{$72.6 \pm 3.6$} &  $76.4 \pm 3.0$ \\

 \textbf{BGCN-NC} & \centering{$71.0 \pm 4.2$}    & \centering{$74.6 \pm 3.3$} & $77.5 \pm 2.4$ \\
 
 \textbf{BGCN-NRWS} & \centering{$\bf{77.7 \pm 2.8}$}    & \centering{$\bf{85.7 \pm 2.4}$} &  $\bf{85.4 \pm 1.1}$ \\
 \hline
\end{tabular}
\vspace{0.2cm}
    \caption{Classification accuracy for Pubmed dataset}
    \label{tab:t3}
\end{table}
\end{center}

\subsection{Hyperparameters}
The hyperparameters used in our experiments are the same as those used for the GCN. We ran the model for 300 epochs with 200 epochs used to pre-train a base GCN classifier. The rest of the 100 epochs use our MCMC graph sampling to minimize loss and increase accuracy. Our base GCN is a $2$-layer network with input dimension of size the number of nodes, $N$, in the dataset, a variational inference layer with input dimension $32$ and output dimension $16$, a hidden layer of dimension $16$, and an output layer of size equal to the number of classes, $K$. We set the learning rate $\alpha$ to a standard $0.01$. To prevent overfitting of our model, we employed dropout regularization with the keep probability set to $0.5$.

\subsection{Results}
We ran a series of experiments with a random split. For the random partition, we randomly sample 5, 10, or 20 labels per class to severely limit the number of classes known for supervision. This is designed to test the limits of the approach under semi-supervision. The random split can give us a more robust measure of the performance of our model.

 The results of each algorithm are based on an average of 50 trial runs with Xavier random weight initialization. We use the accuracy measure, which is simply the number of correctly labeled nodes over the total number of nodes, as our evaluation metric. All the average accuracies for the three datasets are shown in Table \ref{tab:t1}, Table \ref{tab:t2}, and Table \ref{tab:t3}. We observe that our model (BGCN-NRWS) substantially improves node classification results consistently across all levels of supervision and all datasets.


\section{Conclusion}
\label{sec:conclusion}
In this paper, we propose a novel algorithm for Bayesian Graph Convolutional Neural Networks using a neighborhood random walk-based graph sampling method that utilizes graph structure, reduces model overfitting by adding variational inference layers, and enhances the accuracy of the model in semi-supervised node classification. We put heavy constraints on the number of labels randomly sampled per training sample so that we can observe how the model performs under such constraints for semi-supervised learning and we find that our model significantly improves node classification accuracy. Future work includes performing statistical significance tests to determine the level of diversity our model provides. Additionally, more variational layers could be used to approximate posterior quantities effectively, especially as the size of the dataset increases and MCMC methods become computationally expensive.

\section{Acknowledgments}
\label{sec:acknowledgements}
This work was supported in part by National Science Foundation under grant 1946391.

\bibliographystyle{IEEEtran}
\bibliography{Bib}

\begin{thebibliography}{10}
\providecommand{\url}[1]{#1}
\csname url@samestyle\endcsname
\providecommand{\newblock}{\relax}
\providecommand{\bibinfo}[2]{#2}
\providecommand{\BIBentrySTDinterwordspacing}{\spaceskip=0pt\relax}
\providecommand{\BIBentryALTinterwordstretchfactor}{4}
\providecommand{\BIBentryALTinterwordspacing}{\spaceskip=\fontdimen2\font plus
\BIBentryALTinterwordstretchfactor\fontdimen3\font minus
  \fontdimen4\font\relax}
\providecommand{\BIBforeignlanguage}[2]{{%
\expandafter\ifx\csname l@#1\endcsname\relax
\typeout{** WARNING: IEEEtran.bst: No hyphenation pattern has been}%
\typeout{** loaded for the language `#1'. Using the pattern for}%
\typeout{** the default language instead.}%
\else
\language=\csname l@#1\endcsname
\fi
#2}}
\providecommand{\BIBdecl}{\relax}
\BIBdecl

\bibitem{4700287}
F.~{Scarselli}, M.~{Gori}, A.~C. {Tsoi}, M.~{Hagenbuchner}, and
  G.~{Monfardini}, ``The graph neural network model,'' \emph{IEEE Transactions
  on Neural Networks}, vol.~20, no.~1, pp. 61--80, 2009.

\bibitem{inproceedings}
J.~Sun, W.~Guo, D.~Zhang, Y.~Zhang, F.~Regol, Y.~Hu, H.~Guo, R.~Tang, H.~Yuan,
  X.~He, and M.~Coates, ``A framework for recommending accurate and diverse
  items using bayesian graph convolutional neural networks,'' 08 2020, pp.
  2030--2039.

\bibitem{bruna2014spectral}
J.~Bruna, W.~Zaremba, A.~Szlam, and Y.~LeCun, ``Spectral networks and locally
  connected networks on graphs,'' 2014.

\bibitem{henaff2015deep}
M.~Henaff, J.~Bruna, and Y.~LeCun, ``Deep convolutional networks on
  graph-structured data,'' 2015.

\bibitem{duvenaud2015convolutional}
D.~Duvenaud, D.~Maclaurin, J.~Aguilera-Iparraguirre, R.~Gómez-Bombarelli,
  T.~Hirzel, A.~Aspuru-Guzik, and R.~P. Adams, ``Convolutional networks on
  graphs for learning molecular fingerprints,'' 2015.

\bibitem{defferrard2017convolutional}
M.~Defferrard, X.~Bresson, and P.~Vandergheynst, ``Convolutional neural
  networks on graphs with fast localized spectral filtering,'' \emph{Advances
  in Neural Information Processing Systems}, 2017.

\bibitem{kipf2017semisupervised}
T.~N. Kipf and M.~Welling, ``Semi-supervised classification with graph
  convolutional networks,'' in \emph{Proceedings of the 5th International
  Conference on Learning Representations}, 2017.

\bibitem{velickovic2018graph}
P.~Veličković, G.~Cucurull, A.~Casanova, A.~Romero, P.~Liò, and Y.~Bengio,
  ``Graph attention networks,'' \emph{6th International Conference on Learning
  Representations}, 2018.

\bibitem{anirudh2018bootstrapping}
R.~Anirudh and J.~J. Thiagarajan, ``Bootstrapping graph convolutional neural
  networks for autism spectrum disorder classification,'' 2018.

\bibitem{Such_2017}
\BIBentryALTinterwordspacing
F.~P. Such, S.~Sah, M.~A. Dominguez, S.~Pillai, C.~Zhang, A.~Michael, N.~D.
  Cahill, and R.~Ptucha, ``Robust spatial filtering with graph convolutional
  neural networks,'' \emph{IEEE Journal of Selected Topics in Signal
  Processing}, vol.~11, no.~6, p. 884–896, Sep 2017. [Online]. Available:
  \url{http://dx.doi.org/10.1109/JSTSP.2017.2726981}
\BIBentrySTDinterwordspacing

\bibitem{monti2018dualprimal}
F.~Monti, O.~Shchur, A.~Bojchevski, O.~Litany, S.~Günnemann, and M.~M.
  Bronstein, ``Dual-primal graph convolutional networks,'' 2018.

\bibitem{sukhbaatar2016learning}
S.~Sukhbaatar, A.~Szlam, and R.~Fergus, ``Learning multiagent communication
  with backpropagation,'' 2016.

\bibitem{zhang2018bayesian}
Y.~Zhang, S.~Pal, M.~Coates, and D.~Üstebay, ``Bayesian graph convolutional
  neural networks for semi-supervised classification,'' in \emph{Proceedings of
  the AAAI Conference on Artificial Intelligence}, 2018.

\bibitem{pal2019bayesian}
S.~Pal, F.~Regol, and M.~Coates, ``Bayesian graph convolutional neural networks
  using node copying,'' in \emph{Proc. Learning and Reasoning with
  Graph-Structured Representations Workshop (ICML)}, 2019.

\bibitem{chang}
L.~L. Chung~F, D.~TG, and G.~DJ., ``Duplication models for biological
  networks,'' \emph{J. of Computat. Biology}, 2003.

\bibitem{7113345}
R.-H. Li, J.~X. Yu, L.~Qin, R.~Mao, and T.~Jin, ``On random walk based graph
  sampling,'' in \emph{2015 IEEE 31st International Conference on Data
  Engineering}, 2015, pp. 927--938.

\bibitem{JMLR:v15:srivastava14a}
\BIBentryALTinterwordspacing
N.~Srivastava, G.~Hinton, A.~Krizhevsky, I.~Sutskever, and R.~Salakhutdinov,
  ``Dropout: A simple way to prevent neural networks from overfitting,''
  \emph{Journal of Machine Learning Research}, vol.~15, no.~56, pp. 1929--1958,
  2014. [Online]. Available:
  \url{http://jmlr.org/papers/v15/srivastava14a.html}
\BIBentrySTDinterwordspacing

\bibitem{akomand}
A.~Komanduri, ``Improving bayesian graph convolutional networks using markov
  chain monte carlo graph sampling,'' 2021.

\bibitem{jospin2020handson}
L.~V. Jospin, W.~Buntine, F.~Boussaid, H.~Laga, and M.~Bennamoun, ``Hands-on
  bayesian neural networks -- a tutorial for deep learning users,'' 2020.

\bibitem{Goan_2020}
\BIBentryALTinterwordspacing
E.~Goan and C.~Fookes, ``Bayesian neural networks: An introduction and
  survey,'' \emph{Lecture Notes in Mathematics}, p. 45–87, 2020. [Online].
  Available: \url{http://dx.doi.org/10.1007/978-3-030-42553-1_3}
\BIBentrySTDinterwordspacing

\bibitem{gal2016dropout}
Y.~Gal and Z.~Ghahramani, ``Dropout as a bayesian approximation: Representing
  model uncertainty in deep learning,'' 2016.

\bibitem{gal2016bayesian}
Z.~Ghahramani and Y.~Gal, ``Bayesian convolutional neural networks with
  bernoulli approximate variational inference,'' 2016.

\bibitem{hasanzadeh2020bayesian}
A.~Hasanzadeh, E.~Hajiramezanali, S.~Boluki, M.~Zhou, N.~Duffield,
  K.~Narayanan, and X.~Qian, ``Bayesian graph neural networks with adaptive
  connection sampling,'' 2020.

\bibitem{blundell2015weight}
C.~Blundell, J.~Cornebise, K.~Kavukcuoglu, and D.~Wierstra, ``Weight
  uncertainty in neural networks,'' 2015.

\bibitem{hernandezlobato2015probabilistic}
J.~M. Hernández-Lobato and R.~P. Adams, ``Probabilistic backpropagation for
  scalable learning of bayesian neural networks,'' in \emph{Proc. Int. Conf.
  Machine Learning}, 2015.

\bibitem{pmlr-v54-sun17b}
\BIBentryALTinterwordspacing
S.~Sun, C.~Chen, and L.~Carin, ``{Learning Structured Weight Uncertainty in
  Bayesian Neural Networks},'' in \emph{Proceedings of the 20th International
  Conference on Artificial Intelligence and Statistics}, ser. Proceedings of
  Machine Learning Research, A.~Singh and J.~Zhu, Eds., vol.~54.\hskip 1em plus
  0.5em minus 0.4em\relax Fort Lauderdale, FL, USA: PMLR, 20--22 Apr 2017, pp.
  1283--1292. [Online]. Available:
  \url{http://proceedings.mlr.press/v54/sun17b.html}
\BIBentrySTDinterwordspacing

\bibitem{louizos2017multiplicative}
C.~Louizos and M.~Welling, ``Multiplicative normalizing flows for variational
  bayesian neural networks,'' 2017.

\bibitem{Neal}
R.~Neal, ``Bayesian learning via stochastic dynamics,'' in \emph{Advances in
  Neural Information Processing Systems}, S.~Hanson, J.~Cowan, and C.~Giles,
  Eds., vol.~5.\hskip 1em plus 0.5em minus 0.4em\relax Morgan-Kaufmann, 1993.

\bibitem{Sen_Namata_Bilgic_Getoor_Galligher_Eliassi-Rad_2008}
\BIBentryALTinterwordspacing
P.~Sen, G.~Namata, M.~Bilgic, L.~Getoor, B.~Galligher, and T.~Eliassi-Rad,
  ``Collective classification in network data,'' \emph{AI Magazine}, vol.~29,
  no.~3, p.~93, Sep. 2008. [Online]. Available:
  \url{https://ojs.aaai.org/index.php/aimagazine/article/view/2157}
\BIBentrySTDinterwordspacing

\end{thebibliography}

\end{document}